# Machine Learning Applications of Quantum Computing: A Review

**Thien Nguyen, Tuomo Sipola and Jari Hautamäki**

Institute of Information Technology, Jamk University of Applied Sciences, Jyväskylä, Finland

thien.nguyen@student.jamk.fi
tuomo.sipola@jamk.fi
jari.hautamaki@jamk.fi

Abstract: At the intersection of quantum computing and machine learning, this review paper explores the transformative impact these technologies are having on the capabilities of data processing and analysis, far surpassing the bounds of traditional computational methods. Drawing upon an in-depth analysis of 32 seminal papers, this review delves into the interplay between quantum computing and machine learning, focusing on transcending the limitations of classical computing in advanced data processing and applications. This review emphasizes the potential of quantum-enhanced methods in enhancing cybersecurity, a critical sector that stands to benefit significantly from these advancements. The literature review, primarily leveraging Science Direct as an academic database, delves into the transformative effects of quantum technologies on machine learning, drawing insights from a diverse collection of studies and scholarly articles. While the focus is primarily on the growing significance of quantum computing in cybersecurity, the review also acknowledges the promising implications for other sectors as the field matures. Our systematic approach categorizes sources based on quantum machine learning algorithms, applications, challenges, and potential future developments, uncovering that quantum computing is increasingly being implemented in practical machine learning scenarios. The review highlights advancements in quantum-enhanced machine learning algorithms and their potential applications in sectors such as cybersecurity, emphasizing the need for industry-specific solutions while considering ethical and security concerns. By presenting an overview of the current state and projecting future directions, the paper sets a foundation for ongoing research and strategic advancement in quantum machine learning.

## 1. Introduction

Machine learning (ML) is a branch of artificial intelligence (AI). It aims to create systems that can learn from data. Quantum computing (QC) uses the rules of quantum mechanics, which allows it to process information in completely new ways. By combining machine learning with quantum computing, we are laying the groundwork for groundbreaking changes in computer science. As Martín-Guerrero and Lamata (2022) have noted, the synergy of machine learning (ML), quantum computing (QC), and quantum information (QI) is driving the development of Quantum Machine Learning (QML). Recent advances, such as Giuntini et al.'s (2023a) novel quantum-inspired algorithms for classification tasks and Ning et al.'s (2023) quantum approaches to managing large datasets, demonstrate ongoing progress.

One promising application is in cybersecurity, where research such as the exploration of quantum cryptography using continuous-variable quantum neural networks (CV-QNNs) by Shi et al. (2020) demonstrates the potential for robust defenses against cyber threats. However, despite clear progress, challenges such as hardware limitations and algorithm complexity remain. Future research must address these challenges to fully realise the potential of QML in a wide range of applications beyond cybersecurity.

Taking advantage of the recent advances, it is clear that while quantum computing offers a promising avenue for revolutionising machine learning, several gaps and challenges need to be addressed to fully unlock its potential. These observations lead us to formulate the following research questions:

> **RQ1: How do quantum computing principles improve machine learning algorithms?** This question explores the role of quantum computing in improving the performance of machine learning models.

Funded by the European Union.

This work is licensed under Creative Commons Attribution 4.0 International.
To view a copy of this license, visit https://creativecommons.org/licenses/by/4.0/

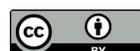 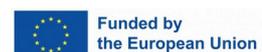

This is the author accepted manuscript version. The original article appeared as: Nguyen, T., Sipola, T., & Hautamäki, J. (2024). Machine Learning Applications of Quantum Computing: A Review. In *Proceedings of the 23rd European Conference on Cyber Warfare and Security (ECCWS)*. Academic Conferences International. https://doi.org/10.34190/eccws.23.1.2258

**RQ2: What are the implications of quantum computing for areas such as cybersecurity?** This question explores how quantum technologies could offer advantages over traditional approaches to securing communication and protecting sensitive data.

**RQ3: How do quantum algorithms compare to classical algorithms in terms of efficiency and application?** This question seeks to understand the comparative advantages and limitations of quantum and classical algorithms for various machine learning tasks.

This review outlines our methodology in Section 2 and examines the current advances in quantum machine learning in the Literature Review in Section 3. Section 4 discusses the key insights and implications of quantum computing in improving machine learning, addressing our research questions (RQ1, RQ2, RQ3). The final section, Section 5, concludes by summarising our findings and suggesting directions for future research. This structured approach aims to provide a clear overview of the role of quantum computing in machine learning, highlighting its potential and the challenges ahead.

## 2. Methodology

This review, part of a larger project exploring applications of quantum computing, examined existing research on its role in machine learning. Following established literature review methods (Levac et al., 2010; Arksey & O'Malley, 2005), we started with a broad search using "Quantum Computing Application" to gather diverse articles. We collected 400 recent publications (2022-2023) from the Science Direct database.

A detailed content analysis identified relevant topics such as AI, machine learning, and applications. This analysis refined the dataset to 287 articles related to quantum computing applications. A further selection was then made using a criterion such as publication date and relationship to the research topic, focusing on articles that explicitly discussed artificial intelligence, resulting in a final set of 32 documents for in-depth review. This methodical approach, based on established methods, provides a comprehensive overview of current trends and developments in quantum machine learning, including its practical applications and future potential.

The methodological approach and detailed stages of the literature review are described, followed by the inclusion of a PRISMA flowchart to visually summarise the process. This diagram (see Figure 1) effectively illustrates the progression from the initial set of 400 documents to the final selection of 32 relevant papers. It highlights the screening, eligibility, and inclusion stages, providing a clear and concise visual representation of the systematic review process, thereby ensuring a thorough selection of the most relevant literature (Page and Moher, 2017).

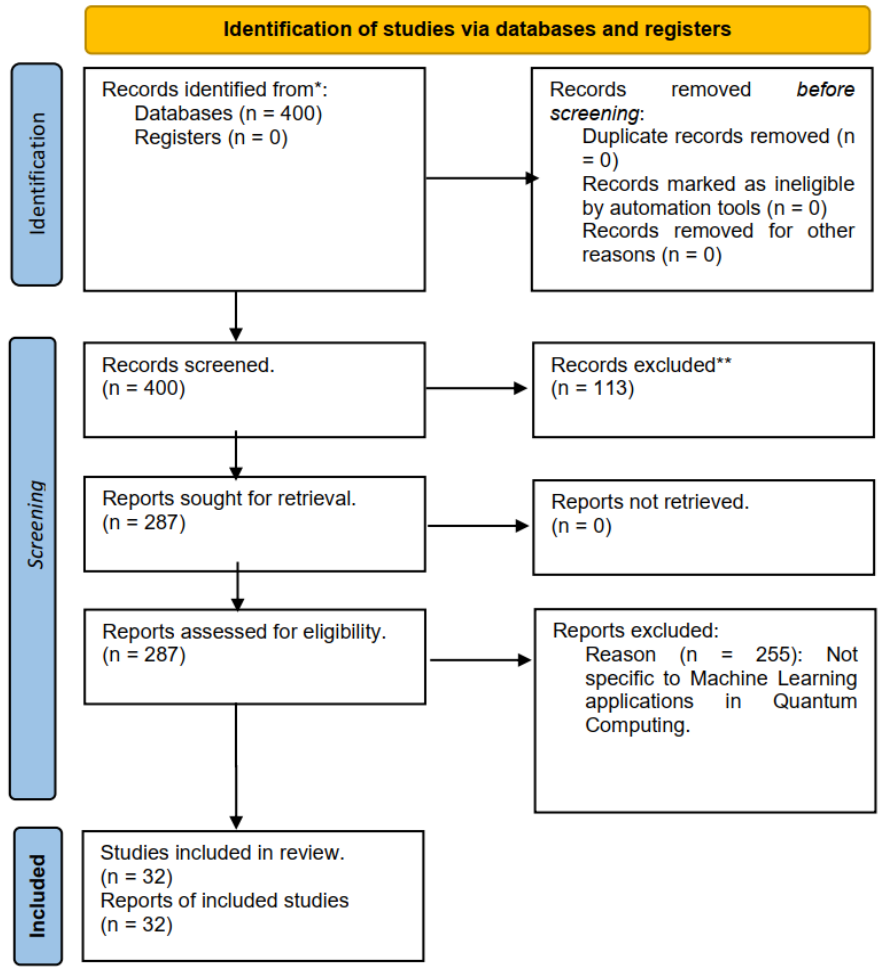

**Figure 1:** PRISMA flow chart diagram showing the systematic selection process of literature from an initial dataset of 400 studies to 32 key papers on machine learning in quantum computing.

## 3. Literature Review

### 3.1 Impact of Quantum Computing on Machine Learning: Key Advances and Algorithm Improvements

The intersection of machine learning (ML) and quantum computing is a burgeoning area of research that is poised to significantly transform data processing and analysis. This review synthesises key studies, highlighting advances in quantum-enhanced computational efficiency and the development of quantum-inspired methods in ML. In particular, the superiority of quantum algorithms in specific applications, especially in pattern recognition and data classification, is demonstrated by the works of Rana et al. (2022) and Houssein et al. (2022). The emergence of quantum neural networks and quantum support vector machines represents a shift towards more advanced quantum computational approaches, as exemplified by Ning, Yang, and Du's (2023) Quantum Kernel Logistic Regression (QKLR), which demonstrates an emerging practical application and transformative potential of quantum computing in ML.

Building on these foundations, Quantum Support Vector Machines (QSVMs), as studied by Zhang et al. (2023) and Rana et al. (2022), exploit quantum computing for more efficient processing of complex, high-dimensional data. QSVMs show a significant improvement over traditional methods in handling data-intensive tasks, highlighting the increasing importance of quantum computing in optimising machine learning models.

Suryotrisongko and Musashi (2022) propose a hybrid quantum-classical deep learning model for botnet detection. While the overall performance is comparable to the classical model, the hybrid model achieves slightly better accuracy (up to 94.7%) in specific cases. The study highlights the sensitivity of the model to initial random seed values and circuit architecture, suggesting the need for further optimisation. Overall, the research represents a promising step towards the application of quantum principles in cybersecurity, but further development is crucial for practical applications.

Exploring the core strengths of quantum machine learning, Tiwari et al. (2023) investigate Quantum Fuzzy Neural Networks (QFNN) for advanced text analysis, including sarcasm detection, while Wei et al. (2023) apply quantum techniques to medical image analysis. Research by Houssein et al. (2022) on a quantum-inspired binary classifier and research by Villalba-Diez et al. (2022) on quantum deep learning further illustrate the progress of the field, highlighting the integration and advancement of existing ideas. Additionally, Martín-Guerrero and Lamata (2022) contribute an in-depth tutorial on various quantum machine learning methods.

Moving to practical implementations, Yulianti et al. (2023) improve ensemble classifiers using a hybrid quantum annealing method, and Li et al. (2023) introduce an innovative quantum approach to k-fold cross-validation, simplifying classification tasks. These developments highlight the impact of quantum computing in improving traditional machine learning techniques.

Further advancing the field, Acampora et al. (2023) propose a novel training method for variational quantum classifiers, addressing key challenges in Noisy-Intermediate Scale Quantum (NISQ) devices. Complementing this, Kim et al. (2023) explore quantum neural networks, in particular quantum convolutional neural networks (QCNNs), demonstrating how quantum and classical computing can be effectively integrated.

Building on this advice, Vadyala and Betgeri (2023) are pioneering the development of Quantum Physics-Informed Neural Networks (PINNs), which combine the reliability of quantum computing with the adaptability of neural networks. Their research advances the computational power of PINNs, particularly in solving complex problems, and represents a significant shift in the design of neural networks, using quantum mechanics to enhance traditional machine learning algorithms.

In the field of quantum-enhanced machine learning, the research of Perkowski (2022) significantly advances the field. His work focuses on areas such as inverse problems, constraint satisfaction, and reversible logic, including the use of Grover quantum oracles. These elements are crucial for the development of sophisticated quantum algorithms to improve machine learning techniques. Perkowski's study (2022) not only enriches the theoretical understanding of quantum ML but also lays the groundwork for its practical applications in various domains.

In exploring the unique capabilities of quantum neural networks, Dong et al. (2022) made a significant contribution by uncovering negational symmetry in Quantum Neural Networks (QNNs) during their study on binary pattern classification. Their study, which focused on binary pattern classification, shows that QNNs exhibit a distinctive behaviour not found in classical neural networks, characterised by an inherent symmetry in the processing of binary patterns and their negation counterparts. This unique behaviour, which differs from classical neural networks, highlights the potential impact of quantum properties on algorithmic behaviour and performance, and provides important observations for the convergence of quantum computing and machine learning.

In a notable development, Wang et al. (2022) introduced a significant advancement in machine learning with their Variational Quantum Extreme Learning Machine (VQELM). This innovative method uses quantum computing to process high-dimensional data more effectively than traditional ML algorithms. Using a unique feature mapping technique for non-linear data, VQELM demonstrates the ability of quantum computing to handle complex datasets and improve the computational efficiency of machine learning.

Collectively, these studies highlight the transformative potential of quantum computing to enhance traditional ML algorithms. They point to a future where quantum principles could fundamentally change the way ML tasks are approached, offering solutions that are more efficient, capable of handling more complex data, and innovatively designed compared to current methods.

## 3.2 Contrasting Approaches

Quantum computing applications in machine learning show a remarkable diversity of approaches, ranging from pure quantum algorithms to innovative hybrid models. This diversity, which is evident in domains ranging from image processing to natural language processing, reflects the evolving landscape of the field. Hybrid quantum-classical models, as applied by researchers such as Villalba-Diez et al. (2022) and Sharma et al. (2023), combine features of both computing paradigms. Their work in improving image analysis and classification processes often outperforms purely classical methods, particularly in tasks requiring high efficiency and accuracy, such as medical image analysis.

Extending this, studies focusing on purely quantum algorithms, such as those by Giuntini et al. (2023a) and Zeng et al. (2023), delve into the development of Quantum Neural Networks (QNNs) and Quantum Support Vector Machines (QSVMs). These studies exploit inherent quantum properties such as superposition and entanglement to enable more efficient processing and analysis of complex, high-dimensional data. This is particularly important in areas such as natural language processing and reinforcement learning, where quantum algorithms show remarkable potential to handle complicated decision-making tasks and to manage large amounts of data more efficiently than their classical counterparts.

In addition, the work of Rana et al. (2022) and Sharma et al. (2023) highlights the advantages of quantum algorithms in specific applications, such as handwriting recognition and text classification. Their results highlight the potential of quantum approaches in a range of diverse machine learning tasks.

Furthermore, in the context of cybersecurity, quantum algorithms have a distinct advantage over their classical counterparts. Studies such as that of Suryotrisongko and Musashi (2022) show how quantum approaches, using quantum mechanics, can process complex data sets more efficiently and provide robust solutions to cybersecurity challenges. This superiority in efficiency, accuracy, and scalability makes quantum algorithms as particularly advantageous for complicated cybersecurity tasks.

In summary, the diverse approaches to quantum machine learning, from pure quantum algorithms to hybrid models, are pushing new frontiers of computational power and grounding quantum principles in practical applications. This range of methods opens up promising avenues for future research, each with the potential to significantly improve data processing and analysis techniques.

## 3.3 Comparative Analyzes

In the field of quantum-enhanced machine learning, comparative studies such as Rana et al.'s work (2022) on Quantum Support Vector Machines (QSVMs) are crucial. They show that QSVM have made significant advances in processing speed and accuracy for handwriting recognition, highlighting their superiority over traditional methods in certain applications. Extending this comparison, Sharma et al. (2023) highlight the effectiveness of quantum kernels in text classification, in particular on how entanglement can improve performance.

Building on these findings, Konar et al. (2023) present a groundbreaking approach to noise-robust image classification with their shallow hybrid classical–quantum spiking feedforward neural network (SQNN). This novel SQNN combines the robustness of classical spiking neural networks (SFNN) with the computational efficiency of quantum circuits, in particular Variational Quantum Circuits (VQC), to significantly improve the performance in processing noisy data. This advance demonstrates the effectiveness of integrating quantum principles into machine learning algorithms, especially in complex, real-world scenarios where data noise is common.

Furthermore, the field of cybersecurity benefits from these comparative analyses. Suryotrisongko and Musashi (2022) show that quantum-enhanced models outperform classical algorithms in identifying and mitigating cyber threats, highlighting the value of quantum approaches in digital security. In industrial applications, Villalba-Diez et al. (2022) use Quantum Deep Learning (QDL) for quality control, achieving faster image processing capabilities than traditional CNNs, demonstrating not only the practical applicability but also the revolutionary potential of quantum computing in traditional ML tasks.

In addition, Houssein et al. (2022) provide a comprehensive overview of various quantum and hybrid models, illustrating a diverse and evolving landscape in quantum machine learning. Complementing this, the integration of quantum principles into image processing, as explored by Singh et al. (2023), provides theoretical insights into quantum improvements in this field. At the same time, Wei et al. (2023) highlight the performance of quantum algorithms in medical image analysis, particularly in managing high-dimensional data more effectively.

### 3.4 Case Studies and Applications

The integration of quantum computing with machine learning is driving innovation in a variety fields. In cybersecurity, the work of Suryotrisongko and Musashi (2022) stands out. They have applied hybrid quantum-classical models to improve botnet DGA detection. This case study is a prime example of how quantum computing can address complex cybersecurity challenges, improving detection accuracy and response times.

Continuing their pioneering efforts, Giuntini et al. (2023a) and Zeng et al. (2023) have introduced quantum-inspired classifiers and hybrid quantum-classical frameworks, respectively, to advance machine learning classification and generative models. Their work illustrates the synergy between quantum and classical computing paradigms in reshaping the practical landscape of quantum computing.

Further illustrating this trend, Villalba-Diez et al. (2022) have applied Quantum Deep Learning (QDL) to industrial quality control, particularly to precision-critical sectors such as the steel industry. Their work demonstrates the practical utility of quantum algorithms in real-world applications. Complementing this, Pandey et al. (2023) have shown how Quantum Machine Learning (QML) can tackle complex natural language processing tasks, such as parts-of-speech tagging in code-mixed datasets, highlighting the versatility of quantum methods for linguistic challenges.

In the field of image generation, Zhou et al. (2023) have made significant progress with a hybrid quantum-classical GAN, optimising quantum generators to improve the efficiency of image generation. This is crucial for data augmentation and image processing, highlighting the broad applicability of quantum ML. In addition, Ovalle-Magallanes et al. (2023) have innovated in quantum convolutional neural networks (QCNNs) by implementing quantum angular encoding with learnable rotation, improving computational performance in image processing tasks and overcoming qubit and circuit depth limitations.

The contributions of Rana et al. (2022) and Wei et al. (2023) also stand out. Rana et al. focus on Quantum Support Vector Machines (QSVM) for image recognition, promoting the superiority of quantum computing in pattern recognition over classical methods. Wei et al. extend this to medical data analysis with Quantum Neural Networks (QNNs), demonstrating their ability to handle high-dimensional medical data.

Exploring quantum-classical hybrid models, Ovalle-Magallanes et al. (2023) introduce an innovative approach to QCNNs that improves their efficiency. Their research, tested on datasets such as MNIST and Fashion-MNIST, shows improved performance and adaptability of these models in image processing, broadening the application scope of quantum computing in machine learning.

Taken together, these case studies not only showcase the practical and theoretical applications of quantum computing in machine learning but also highlight the evolving and diverse nature of the field. They illustrate how quantum computing is able to enhance traditional methods and open up new possibilities in data processing and analysis, thereby underlining its potential to deliver significant advances across a range of industries.

### 3.5 Future Trends and Research Directions

The findings suggest a trajectory towards more integrated and advanced quantum machine learning systems. Pioneering work in cybersecurity, such as the work by Suryotrisongko and Musashi (2022), indicates a growing trend towards using quantum computing to improve digital security measures. Future research is likely to focus on further optimising quantum algorithms for broader applications, including more robust cybersecurity solutions. In addition, the development of quantum hardware and the exploration of new quantum-classical hybrid models, as highlighted by Villalba-Diez et al. (2022) and Wei et al. (2023), are expected to be key areas of development. These advances will not only enhance the capabilities of quantum machine learning in complex data processing and threat detection but also broaden its application in various fields.

Building on this perspective, Amato et al. (2023) present 'QuantuMoonLight,' a user-friendly, low-code platform designed to streamline quantum machine learning experiments. This platform facilitates the comparison of quantum algorithms with classical ones, making quantum technologies more accessible and encouraging community collaboration. 'QuantuMoonLight' represents a significant advance in quantum machine learning, broadening its appeal and understanding.

Following this trend, Liu et al. (2023) make significant advances in quantum reinforcement learning, applying quantum algorithms to complex challenges such as the Multi-Arm Bandit and Grid problems. Their research, which demonstrates improved efficiency and speed over traditional methods, demonstrates the strengths of quantum computing in dealing with large, complicated issues, and sets the stage for more advanced quantum learning models.

In addition, Molteni et al. (2023) make a significant contribution to quantum echo-state networks by optimising their memory reset rate, particularly for time-sequential tasks. By demonstrating enhanced performance on IBM quantum hardware, their work underscores the importance of quantum reservoir computing in Quantum Machine Learning (QML). This advance in memory scalability and efficiency enriches our understanding of the potential of quantum machine learning algorithms in complex data processing.

A comprehensive analysis by Jadhav, Rasool, and Gyanchandani (2023) explores the broad potential of Quantum Machine Learning (QML) in various fields. They highlight advances in QML algorithms and their real-world applications, particularly in the efficient management of large datasets using quantum principles. The review also discusses strategies for integrating classical and quantum approaches and illustrates the effectiveness of QML in areas such as pattern recognition and bioinformatics.

Next, Kwak et al. (2023) look at Quantum Distributed Deep Learning (QDDL), assessing its potential to improve data security and computational efficiency. They evaluate different QDDL architectures, emphasising their ability to provide secure data management and quantum communication protocols. This research marks an important step in the application of quantum principles to deep learning, addressing both efficiency and security challenges.

Finally, Chen, Samuel, and Yen-Chi (2023) explore a novel asynchronous training method in Quantum Reinforcement Learning (QRL). Focusing on advanced actor-critic quantum policies, their work shows that asynchronous training can rival or surpass classical methods in efficiency, simplifying training and extending the applications of QRL in complex tasks. This suggests a promising direction for more effective quantum machine learning algorithms.

### 3.6 Limitations and Challenges

Despite significant progress, the application of quantum computing in machine learning, and cybersecurity in particular, faces notable challenges. As highlighted by Suryotrisongko and Musashi (2022), data privacy concerns and the complexity of quantum algorithms are significant hurdles in security applications. These issues are compounded by the limitations of current quantum hardware and scalability challenges. Studies such as those by Tiwari et al. (2023) and Zeng et al. (2023) also highlight the difficulties of integrating complex quantum algorithms. Overcoming these challenges will be critical to the practical application and effectiveness of quantum ML techniques in cybersecurity and other domains.

### 4. Discussion

Our exploration of Quantum Machine Learning (QML) reveals a vibrant field where advances in quantum computing are being used to improve machine learning capabilities. As detailed in Table 1, breakthroughs like Quantum Kernel Logistic Regression (QKLR) by Ning et al. (2023) aim to overcome limitations of traditional logistic regression in handling complex data, promising improved pattern recognition and data classification. Similarly, Quantum Support Vector Machines (QSVMs) show how efficiency gains in processing complex data (Rana et al., 2022; Zhang et al., 2023), although further research is needed. Hybrid Quantum-Classical Deep Learning models, as exemplified by the work of Suryotrisongko and Musashi (2022), achieve comparable or even superior accuracy in areas such as cybersecurity applications, although they require optimisation due to sensitivity to initial conditions.

Table 2 delves into the practical applications of these advances, showing real-world examples such as the significant improvement in botnet detection accuracy achieved by Suryotrisongko and Musashi's (2022) hybrid model in the cybersecurity domain. Quantum techniques, as explored by Wei et al. (2023) demonstrate effectiveness in advanced medical image analysis, while Villalba-Diez et al. (2022) showcase the potential of Quantum Deep Learning (QDL) for enhanced image processing capabilities in industrial quality control. These diverse applications highlight the transformative potential of QML across different sectors.

However, as shown in Table 3, challenges remain. Limited hardware capabilities, characterised by the limited number of qubits (quantum bits) and susceptibility to error of current quantum computing devices, limit the complexity and scalability of QML applications. As quantum computers become more powerful, the question of how to protect sensitive data becomes even more pressing. Addressing these concerns requires the development of new, quantum-resistant encryption techniques.

**Table 1:** Key Advancements in Quantum Machine Learning

| Advancement | Authors | Impact | Challenges |
| --- | --- | --- | --- |
| Quantum Kernel Logistic Regression (QKLR) | Ning et al. (2023) | Enhances pattern recognition and data classification | Limited to linear problems. (Logistic regression is a weakness, hence development of QKLR). |
| Quantum Support Vector Machines (QSVMs) | Rana et al. (2022), Zhang et al. (2023) | Improves efficiency in processing complex data | Need for more comprehensive exploration |
| Hybrid Quantum-Classical Deep Learning Models | Suryotrisongko, Musashi (2022) | Comparable or superior accuracy in cybersecurity applications | Sensitivity to initial conditions; optimization required |

**Table 2:** Case Studies and Applications

| Application Domain | Study | Model/Technique | Key Findings |
| --- | --- | --- | --- |
| Cybersecurity | Suryotrisongko, Musashi (2022 | Hybrid Quantum-Classical Model | Achieved up to 94.7% accuracy in botnet detection |
| Medical Image Analysis | Wei et al. (2023) | Quantum Techniques | Demonstrated effectiveness in advanced analysis |
| Industrial Quality Control | Villalba-Diez et al. (2022) | Quantum Deep Learning (QDL) | Enhanced image processing capabilities |

**Table 3:** Challenges and Limitations

| Challenge | Description | Implications for QML |
| --- | --- | --- |
| Limited Hardware Capabilities | The current quantum computing hardware offers limited qubits and is prone to errors. | Constrains the complexity and scalability of QML applications. |
| Data Privacy Concerns | The capacity of quantum computing to disrupt present encryption practices introduces concerns regarding data security. | Necessitates the development of new, quantum-resistant encryption techniques. |

## 5. Conclusion

Our review explored the current state of Quantum Machine Learning (QML), focusing on advances in algorithms (such as Quantum Kernel Logistic Regression and Quantum Support Vector Machines), potential applications in various fields (from strengthening cybersecurity to pioneering medical image analysis and improving industrial quality control), and existing challenges (hardware limitations and privacy concerns). These advances, which address our first research question (RQ1), show significant potential for improved pattern recognition and complex data processing in machine learning. The potential applications (RQ2) highlight the transformative potential of QML, but the challenges (RQ3) represent critical barriers to widespread adoption. While this review provides a quantitative overview, it highlights the need for a deeper exploration to bridge the gap between theory and practice, paving the way for QML to unlock its full potential and revolutionise the future of machine learning.

**Acknowledgements**

We are grateful to Emils Bagirovs, Grigory Provodin and Ummar Ahmed for their help with data collection and assistance with database creation. This research was partially supported by the ResilMesh project, funded by the European Union's Horizon Europe Framework Programme (HORIZON) under grant agreement 101119681. The authors would like to thank Ms. Tuula Kotikoski for proofreading the manuscript.

**References**

Acampora, G., Chiatto, A. and Vitiello, A., 2023. Training circuit-based quantum classifiers through memetic algorithms. Pattern Recognition Letters, 170, p.32-38. https://doi.org/10.1016/j.patrec.2023.04.008

Amato, F., Cicalese, M., Contrasto, L., Cubicciotti, G., D'Ambola, G., La Marca, A., Pagano, G., Tomeo, F., Robertazzi, G. A., Vassallo, G., Acampora, G., Vitiello, A., Catolino, G., Giordano, G., Lambiase, S., Pontillo, V., Sellitto, G., Ferrucci, F., and Palomba, F. (2023). QuantuMoonLight: A low-code platform to experiment with quantum machine learning. *SoftwareX, 22, 101399.* https://doi.org/10.1016/j.softx.2023.101399

Chen, S. Y.-C., et al. (2023). Asynchronous training of quantum reinforcement learning. *Procedia Computer Science, 222(2023), 321–330.* https://doi.org/10.1016/j.procs.2023.08.171

Dong, N., Kampffmeyer, M., Voiculescu, I., and Xing, E. (2022). Negational symmetry of quantum neural networks for binary pattern classification. *Pattern Recognition, 129(2022), 108750.* https://doi.org/10.1016/j.patcog.2022.108750

D. Levac, H. Colquhoun, and K. K. O'Brien, "Scoping studies: advancing the methodology," Implementation Science, vol. 5, 2010.

Giuntini, R., Holik, F., Park, D.K., Freytes, H., Blank, C., & Sergioli, G. (2023a). Quantum-inspired algorithm for direct multi-class classification. *Applied Soft Computing*, 134, 109956. https://doi.org/10.1016/j.asoc.2022.109956

Giuntini, R., Granda Arango, A.C., Freytes, H., Holik, F.H., & Sergioli, G. (2023b). Multi-class classification based on quantum state discrimination. Fuzzy Sets and Systems, 467, pp. 108509. https://doi.org/10.1016/j.fss.2023.03.012

H. Arksey and L. O'Malley, "Scoping studies: towards a methodological framework," International Journal of Social Research Methodology, vol. 8, pp. 19–32, 2005.

Houssein, E.H., et al., 2022. Machine learning in the quantum realm: The state-of-the-art challenges and future vision. *Expert Systems With Applications, 194, p.116512.* https://doi.org/10.1016/j.eswa.2022.116512

Jadhav, A., Rasool, A., and Gyanchandani, M. (2023). Quantum Machine Learning: Scope for Real-World Problems. *Procedia Computer Science, 218, 2612–2625.* https://doi.org/10.1016/j.procs.2023.01.235

Kim, J., Huh, J. and Park, D.K., 2023. Classical-to-quantum convolutional neural network transfer learning. *Neurocomputing, 555, pp.126-643.* https://doi.org/10.1016/j.neucom.2023.126643

Konar, D., Sarma, A. D., Bhandary, S., Bhattacharyya, S., Cangi, A., and Aggarwal, V. (2023). A shallow hybrid classical–quantum spiking feedforward neural network for noise-robust image classification. *Applied Soft Computing, 136, 110099*. https://doi.org/10.1016/j.asoc.2023.110099

Kwak, Y., Yun, W. J., Kim, J. P., Cho, H., Park, J., Choi, M., Jung, S., and Kim, J. (2023). Quantum distributed deep learning architectures: Models discussions and applications. ICT Express, 9(1), 486–491. https://doi.org/10.1016/j.icte.2022.08.004

Li, J., Gao, F., Lin, S., Guo, M., Li, Y., Liu, H., Qin, S., and Wen, Q. (2023). Quantum k-fold cross-validation for nearest neighbor classification algorithm. *Physica A: Statistical Mechanics and its Applications, 611, 128435*. https://doi.org/10.1016/j.physa.2022.128435

Liu, Y.-P., Jia, Q.-S., and Wang, X. (2022). Quantum reinforcement learning method and application based on value function. *IFAC-PapersOnLine, 55(11), 132–137.* https://doi.org/10.1016/j.ifacol.2022.08.061

Martín-Guerrero, J. D., and Lamata, L. (2022). Quantum Machine Learning: A tutorial. *Neurocomputing, 470, 457-461.* https://doi.org/10.1016/j.neucom.2021.02.102

Molteni, R., Destri, C. and Prati, E., 2023. Optimization of the memory reset rate of a quantum echo-state network for time sequential tasks. *Physics Letters A, 465, p.128713.* https://doi.org/10.1016/j.physleta.2023.128713

Ning, T., Yang, Y. and Du, Z., 2023. Quantum kernel logistic regression-based Newton method. *Physica A: Statistical Mechanics and its Applications, 611, p.128454.* https://doi.org/10.1016/j.physa.2023.128454

Ovalle-Magallanes, E., Alvarado-Carrillo, D.E., Avina-Cervantes, J.G., Cruz-Aceves, I., and Ruiz-Pinales, J. (2023). Quantum angle encoding with learnable rotation applied to quantum–classical convolutional neural networks. *Applied Soft Computing, 141, 110307*. https://doi.org/10.1016/j.asoc.2023.110307

Page, M.J. and Moher, D., 2017. Evaluations of the uptake and impact of the Preferred Reporting Items for Systematic reviews and Meta-Analyses (PRISMA) Statement and extensions: a scoping review.

*Systematic Reviews, 6(1), p.263*. https://doi.org/10.1186/s13643-017-0663-8

Pandey, S., Basisth, N. J., Sachan, T., Kumari, N., and Pakray, P. (2023). Quantum Machine Learning for Natural Language Processing Applications. *Physica A, 627(2023), 129123.* https://doi.org/10.1016/j.physa.2023.129123

Perkowski, M. (2022). Inverse problems constraint satisfaction reversible logic invertible logic and Grover quantum oracles for practical problems. *Science of Computer Programming, 218(2022), 102775.* https://doi.org/10.1016/j.scico.2022.102775

Rana, A., Vaidya, P. and Gupta, G., 2022. A comparative study of quantum support vector machine algorithm for handwritten recognition with support vector machine algorithm. *Materials Today: Proceedings, 56, pp.2025-2030.* https://doi.org/10.1016/j.matpr.2021.11.350

Sharma, D., Singh, P., and Kumar, A. (2023). The role of entanglement for enhancing the efficiency of quantum kernels towards classification. *Physica A, 625, 128938.* https://doi.org/10.1016/j.physa.2023.128938

Shi, J., Chen, S., Lu, Y., Feng, Y., Shi, R., Yang, Y., and Li, J. (2020). An Approach to Cryptography Based on Continuous-Variable Quantum Neural Network. *Scientific Reports, natureresearch.* https://www.nature.com/articles/s41598-020-58928-1

Singh, S., Pandian, M. T., Aggarwal, A. K., Awasthi, S. P., Bhardwaj, H., and Pruthi, J. (2023). Quantum learning theory: A classical perspective for quantum image. *Materials Today: Proceedings, 80(2023), 2786–2793.* https://doi.org/10.1016/j.matpr.2021.07.039

Suryotrisongko, H., and Musashi, Y. (2022). Evaluating Hybrid Quantum-Classical Deep Learning for Cybersecurity Botnet DGA Detection. *Procedia Computer Science, 197(2022), 223–229.* https://doi.org/10.1016/j.procs.2021.12.135

Tiwari, P., Zhang, L., Qu, Z., and Muhammad, G. (2023). Quantum Fuzzy Neural Network for multimodal sentiment and sarcasm detection. *Information Fusion, 103(2024), 102085.* https://doi.org/10.1016/j.inffus.2023.102085

Vadyala, S. R., and Betgeri, S. N. (2023). General implementation of quantum physics-informed neural networks. *Array, 18, 100287*. https://doi.org/10.1016/j.array.2023.100287

Villalba-Diez, J., Ordieres-Mere, J., Gonzalez-Marcos, A., and Soto Larzabal, A. (2022). Quantum deep learning for steel industry computer vision quality control. *IFAC PapersOnLine, 55(2), 337–342.* https://doi.org/10.1016/j.ifacol.2022.04.216

Wang, Y., Lin, K.-Y., Cheng, S., and Li, L. (2022). Variational quantum extreme learning machine. *Neurocomputing, 512, 83–99.* https://doi.org/10.1016/j.neucom.2022.09.068

Wei, L., Liu, H., Xu, J., Shi, L., Shan, Z., Zhao, B., and Gao, Y. (2023). Quantum machine learning in medical image analysis: A survey. *Neurocomputing, 525, 42–53.* https://doi.org/10.1016/j.neucom.2023.01.049

Yulianti, L.P., et al., 2023. A hybrid quantum annealing method for generating ensemble classifiers. *Journal of King Saud University - Computer and Information Sciences, 35, p.101831.* https://doi.org/10.1016/j.jksuci.2023.101831

Zeng, Q.-W., Ge, H.-Y., Gong, C., Zhou, N.-R., et al. (2023). Conditional quantum circuit Born machine based on a hybrid quantum–classical framework. *Physica A: Statistical Mechanics and its Applications, 618, 128693.* https://doi.org/10.1016/j.physa.2023.128693

Zhang, R., Wang, J., Jiang, N., and Wang, Z. (2023). Quantum support vector machine without iteration. *Information Sciences, 635, 25–41.* https://doi.org/10.1016/j.ins.2023.03.106

Zhou, N.-R., Zhang, T.-F., Xie, X.-W., Wu, J.-Y. (2023). Hybrid quantum–classical generative adversarial networks for image generation via learning discrete distribution. *Signal Processing: Image Communication, 110, 116891.* https://doi.org/10.1016/j.image.2022.116891